\documentclass[letterpaper]{article} % DO NOT CHANGE THIS
\usepackage{aaai25}  % DO NOT CHANGE THIS
\usepackage{times}  % DO NOT CHANGE THIS
\usepackage{helvet}  % DO NOT CHANGE THIS
\usepackage{courier}  % DO NOT CHANGE THIS
\usepackage[hyphens]{url}  % DO NOT CHANGE THIS
\usepackage{graphicx} % DO NOT CHANGE THIS
\usepackage{natbib}  % DO NOT CHANGE THIS AND DO NOT ADD ANY OPTIONS TO IT
\usepackage{caption} % DO NOT CHANGE THIS AND DO NOT ADD ANY OPTIONS TO IT
\usepackage{algorithm}
\usepackage{algorithmic}

\usepackage{threeparttable}
\usepackage{multirow}
\usepackage{booktabs}
\usepackage{pifont}
\usepackage{subfigure}
\usepackage{newfloat}
\usepackage{listings}
\usepackage{amssymb}
\usepackage{amsmath}
\newcommand{\cmark}{\text{\ding{51}}}

\DeclareCaptionStyle{ruled}{labelfont=normalfont,labelsep=colon,strut=off} % DO NOT CHANGE THIS
\floatstyle{ruled}
\newfloat{listing}{tb}{lst}{}
\floatname{listing}{Listing}
\title{TAMER: Tree-Aware Transformer for Handwritten Mathematical Expression Recognition}
\author{
    Jianhua Zhu\textsuperscript{\rm 1},
    Wenqi Zhao\textsuperscript{\rm 1} ,
    Yu Li\textsuperscript{\rm 1} ,
    Xingjian Hu\textsuperscript{\rm 1},
    Liangcai Gao\textsuperscript{\rm 1}\thanks{Corresponding author.}\\
}
\affiliations{
    \textsuperscript{\rm 1}Wangxuan Institute of Computer Technology, Peking University, Beijing, China\\

    zhujianhuapku@pku.edu.cn, \{wenqizhao, liyu\}@stu.pku.edu.cn,  \{huxingjian, gaoliangcai\}@pku.edu.cn
    
}

\usepackage{bibentry}
\begin{document}

\maketitle

\begin{abstract}
Handwritten Mathematical Expression Recognition (HMER) has extensive applications in automated  grading and office automation. However, existing sequence-based decoding methods, which directly predict \LaTeX\ sequences, struggle to understand and model the inherent tree structure of \LaTeX\ and often fail to ensure syntactic correctness in the decoded results. To address these challenges, we propose a novel model named TAMER (Tree-Aware Transformer) for handwritten mathematical expression recognition. TAMER  introduces an innovative Tree-aware Module while maintaining the flexibility and efficient training of Transformer. TAMER combines the advantages of both sequence decoding and tree decoding models by jointly optimizing sequence prediction and tree structure prediction tasks, which enhances the model's understanding and generalization of complex mathematical expression structures. During inference, TAMER  employs a Tree Structure Prediction Scoring Mechanism to improve the structural validity of the generated \LaTeX\ sequences. Experimental results on CROHME datasets demonstrate that TAMER outperforms traditional sequence decoding and tree decoding models, especially in handling complex mathematical structures, achieving state-of-the-art (SOTA) performance.
\end{abstract}

\begin{links}
    \link{Code}{https://github.com/qingzhenduyu/TAMER/}
    % \link{Datasets}{https://aaai.org/example/datasets}
    % \link{Extended version}{https://aaai.org/example/extended-version}
\end{links}

\section{Introduction}
The task of  Handwritten Mathematical Expression Recognition(HMER) holds a unique and important place in the field of Optical Character Recognition (OCR). Compared to traditional OCR tasks, it faces more complex challenges. Traditional character recognition models typically focus on recognizing linearly arranged text sequences, whereas the structure of handwritten mathematical expressions is not one-dimensional and linear but exhibits multi-layered and highly structured characteristics. For example, mathematical symbols such as fractions, square roots, and integrals involve not only basic characters but also complex structures like subscripts, superscripts, and nested symbols, all of which greatly increase the difficulty of recognition. Therefore, a core challenge in recognizing handwritten mathematical expressions is how to effectively encode the structural information of the expressions within the model. This requires the model to not only accurately identify characters in the image but more importantly, to construct the complex relationships between these characters. Enhancing the model's understanding and inference abilities regarding the structure of mathematical expressions, enabling it to correctly decode mathematical expressions that follow mathematical syntax rules, remains an important research question that the field continues to explore.
\begin{figure}[t]
\centering
\includegraphics[width=0.95\columnwidth]{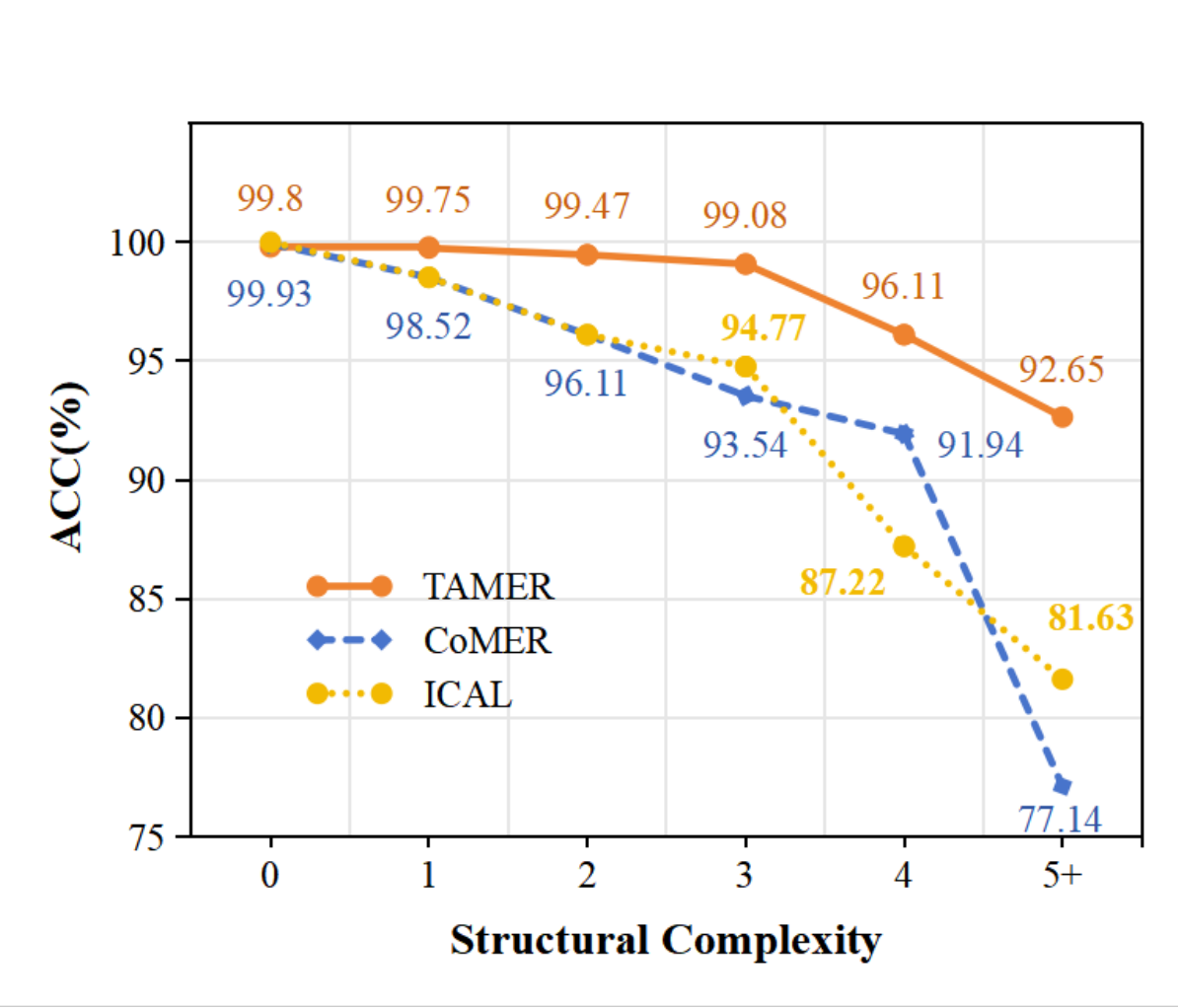}
\caption{\textbf{Bracket Matching Accuracy under different structural complexities on CROHME 2014(in \%).} TAMER maintains a bracket matching accuracy of over 92\% across all levels of structural complexity, significantly outperforming CoMER and ICAL.} 
\label{figure:bracket_acc}
\end{figure}
The representation of mathematical expressions is mainly categorized into two types: sequence-based and tree-structure-based representations. The former is typified by \LaTeX\ expressions, which convert mathematical expressions into one-dimensional character sequences and use special structural control symbols (such as left curly braces ``\verb|{|" and right curly braces ``\verb|}|") to express the hierarchical structure of the expressions. For example, the mathematical expression $x^2$ is represented in \LaTeX\ as ``\verb|x^{2}|". The tree-structure-based representation, on the other hand, decomposes mathematical expressions into tuples, each tuple $(p, c, r)$ comprising a parent node $p$, a child node $c$, and the relationship $r$ between them. For instance, the same mathematical expression $x^2$ would be represented in this method as $(x, 2, \text{sup})$. Based on these two methods of representing mathematical expressions, the field of HMER has also developed two corresponding decoding methods: sequence-based decoding methods and tree-based decoding methods.

Sequence-based decoding methods target \LaTeX\ sequences, which have the advantage of being a widely used method for expressing mathematical expressions. This compatibility not only aligns well with existing research in fields such as NLP and OCR but also offers strong versatility. However, this approach has its limitations. First, during the training process, the tree-like structural characteristics of mathematical expressions are not fully considered, which may lead to insufficient understanding of expressions structures by the model, thus affecting its generalization ability when dealing with complex structures. Second, as targets for sequence prediction, \LaTeX\ sequences do not ensure the grammatical correctness of the model's decoding results, especially when training on a relative small dataset, like CROHME dataset. For example, when dealing with complex mathematical expressions, the model might generate mismatched left and right curly braces, leading to predictions that do not conform to \LaTeX\ syntax standards.

Tree-structure-based decoding methods treat mathematical expressions as tree-like structures for prediction and decoding. The advantage of this approach is that the model design fully integrates the characteristics of tree structures, theoretically enhancing the model's ability to understand mathematical expressions and ensuring that it always decodes valid expressions during the prediction and inference process. Despite these benefits, tree-structure decoding models have their limitations. First, these models rely on RNN for construction, which cannot fully utilize the efficient parallel training features of Transformer models. Additionally, their decoding method is relatively complex and lacks the universality and broad applicability compared to \LaTeX\ expressions. Most importantly, in practical experiments, the performance of tree-structure decoding models is often inferior to that of sequence decoding models, leading to a lack of widespread adoption of tree-structure decoding methods. Therefore, in practical engineering applications, engineers often prefer to choose sequence decoding models.

Addressing the respective advantages and drawbacks of these two decoding methods, in this work, we integrate the strengths of both sequence-based decoding and tree-based decoding models to propose a novel HMER model that is aware of tree structures of expressions, named TAMER (Tree-Aware Transformer). Our contributions are as follows:

(1) Without harming the flexibility and efficient training efficiency of sequence decoding models, TAMER introduces a novel Tree-Aware Module(TAM). It jointly optimizes both sequence prediction and tree structure prediction tasks, enhancing the model's understanding and generalization capabilities for complex expression structures.

(2) In the inference phase, TAMER integrates a Tree Structure Prediction Scoring Mechanism into the beam search algorithm, allowing the model to consider the rationality of the expression tree structure while decoding \LaTeX\ sequences. This mechanism ensures that the model's recognition results are more complete and accurate in terms of grammatical structure.

(3) Experimental results across CROHME datasets demonstrate that the TAMER outperforms traditional sequence-based decoding and tree-based decoding methods, highlighting the potential of combining these two decoding methods. TAMER achieves expression recognition rate (ExpRate) of 61.23\%, 60.26\%, and 61.97\% on the CROHME 2014/2016/2019 test sets~\cite{mouchere2014icfhr, mouchere2016icfhr2016, mahdavi2019icdar}, respectively. Additionally, TAMER excels in avoiding syntactical errors, particularly in bracket matching, significantly outperforming other methods—especially on the CROHME datasets, which have relatively small training sets.
\section{Related Work}
% 翻译下面这段话
% 手写公式识别,由于手写公式具有嵌套层次结构、书写风格多样等特点,一直是OCR领域的一个研究难点。在传统方法中,手写数学公式识别通常包含符号识别(Symbol Recognition)和结构分析(Structural  Analysis)两个步骤。
% 随着深度学习的崛起，该方法已被广泛应用于多种光学字符识别(OCR)任务,并且性能大大超过了传统方法。在手写数学公式识别领域,基于深度学习的方法也已成为主流,  这些方法通常采用编码器-解码器(Encoder-Decoder)  架构来构建模型,根据解码方式的不同,深度学习方法可分为基于序列解码的方法和基于树结构的方法
HMER has long been a challenging research area within Optical Character Recognition (OCR) due to the nested hierarchical structures and diverse writing styles of handwritten expressions. Traditional HMER methods \cite{599029,ha1995understanding,alvaro2014recognition,kosmala1999line,winkler1996hmm,chan1998elastic,vuong2010towards,chou1989recognition,sakai1961syntax,chan2000efficient,zanibbi2002recognizing,toyota2006structural} typically involve two steps: symbol recognition and structural analysis.
% With the rise of deep learning, this approach has been widely adopted for various OCR tasks, significantly surpassing traditional methods\cite{alzubaidi2021review}. 
Recently, deep learning methods have become predominant. These methods commonly employ an encoder-decoder architecture. Depending on the difference of decoding manner, deep learning methods can be classified into sequence-based decoding methods and tree-based decoding methods.

\subsection{Sequence-based decoding methods}

\subsubsection{RNN-based Models} In 2017, the WAP (Watch, Attend, and Parse) \cite{zhang2017watch} pioneered the use of deep neural networks for HMER, initiating significant advancements in this research field.
Following WAP, DenseWAP \cite{zhang2018multi} replace VGG encoder with DenseNet. Subsequent research has frequently adopted DenseNet as the primary backbone network for the visual encoder.
PAL\cite{wu2019palv1} and PALv2\cite{wu2020palv2} introduces an adversarial learning method utilizing handwritten-printed sample pairs. By learning semantic-invariant features, this approach enhances the accuracy of recognizing \LaTeX \ expressions in various styles
% PAL和 PALv2 引入了一种利用手写-打印样本对的对抗学习方法。这种方法通过学习语义不变特征，提高了识别各种风格的 LaTeX 公式的准确性。
The ABM\cite{bian2022handwritten} incorporates mutual learning loss between two directions to to mitigate the issue of output imbalance.
Recently, the CAN \cite{li2022counting} incorporates a Multi-Scale Counting Module that employs a symbolic counting task as an auxiliary task, which is jointly optimized alongside the expression sequence prediction task. 
All of these models employ decoders are based on RNN. However, existing research \cite{bengio1993problem} indicates that models based on RNN have inherent limitations in modeling long-range dependencies between characters.

\subsubsection{Transformer-based Models} In the field of natural language processing, Transformer models have progressively supplanted traditional RNN architectures\cite{he2016deep,devlin2018bert}.
BTTR \cite{zhao2021handwritten}is the first to employ Transformer decoder for HMER and introduces a bidirectional training strategy to address the issue of output imbalance while fully leveraging bidirectional language information.
% ABM 模型进一步增加了两个方向间的互学习损失，并使用单向推理策略有效提升了推理预测的效率。
% CoMER设计了Attention Refinement Module，将用于RNN的coverage attention 引入到Transformer解码器中，缓解了缺乏覆盖问题.
% 没有影响到并行训练。CoMER也成为后续基于Transformer解码器的重要backbone. 
CoMER \cite{zhao2022comer} designs an Attention Refinement Module(ARM) that integrates coverage attention mechanism from RNN into the Transformer decoder, thereby alleviating the lack of coverage problem.
Moreover, CoMER maintains the parallel training capabilities of the Transformer model and achieves impressive experimental results, making it an important backbone for subsequent methods.
Based on CoMER and CAN, the GCN\cite{zhang2023general} manually categorizes mathematical symbols into broader groups and utilizes the General Category Recognition Task as a supplementary task for joint optimization with the HMER task.
The ICAL\cite{zhu2024ical} introduces the Implicit Character Construction Module (ICCM) to model implicit character information within the \LaTeX\ sequence. Additionally, it utilizes a Fusion Module to integrate ICCM outputs, thereby correcting the predictions of the Transformer Decoder.

Different from the autoregressive sequence-based decoding methods developed in recent years, NAMER \cite{liu2024namer} introduces a non-autoregressive modeling framework, which applies a parallel graph decoder to revise predicted visible symbols and establish connectivity between them, significantly reducing memory usage and inference costs.

\subsection{Tree-based decoding methods}

% In 2020, the DenseWAP-TD \cite{zhang2020treedecoder}, marking the first use of a tree decoding model in HMER. 
DenseWAP-TD \cite{zhang2020treedecoder} is the first work to employ a tree-based decoding model in the field of HMER, replacing the GRU decoder that directly regresses the \LaTeX\ sequence with a decoder based on a two-dimensional tree structure. 
% Specifically, the decoder component of the model consists of three parts: the parent decoder, the child decoder, and the relation classifier. It jointly considers the parent and child nodes to calculate the probability distribution of their relationship, ultimately decoding to obtain the subtree sequence. This model demonstrates enhanced generalization ability when handling mathematical expressions with complex nested structures.  
The TDv2 \cite{wu2022tdv2} is primarily optimized for the tree-based decoding technology in the DenseWAP-TD , with a particular focus on the diverse utilization of tree structures and the simplification of the decoding process. Furthermore, during the training process of the TDv2 model, various conversion methods are applied to the same \LaTeX\  string. The Syntax-Aware Network (SAN)\cite{yuan2022syntax} converts \LaTeX\ sequences into a parsing tree, employing syntactic rules to reformulate the prediction of \LaTeX\ sequences as a tree traversal task. In addition, SAN incorporates a novel Syntax-Aware Attention Module, effectively integrating syntactic information into the recognition process. 
% However, most tree-based decoding methods have several limitations. They depend on RNNs, which restrict the efficient parallel training capabilities of transformers and lead to poor performance relative to sequence-based decoding methods. Furthermore, their decoding process tends to be more intricate and less versatile compared to \LaTeX\  expressions.
% This approach reduces contextual dependencies and enhances the generalization capabilities of the decoder.
% add  disadvantange ,and introduce SAN but comment
% However, both models utilize a character-by-character prediction decoding paradigm and do not explicitly account for the syntactic structure of mathematical expressions during the learning process. This approach may lead to generated tree structures that lack reasonable syntactic constraints.
 
% Syntax-Aware Network (SAN)\cite{yuan2022syntax} exploits detailed grammar rules and syntactic information to efficiently partition a syntax tree into distinct components, thereby reducing prediction errors in the structure of mathematical expressions.
% The SAN model [28] converts the LATEX sequence into a parsing tree and designs a series of syntactic rules to transform the problem of predicting LATEX sequences into a tree traversal process. Additionally, SAN introduces a new Syntax-Aware Attention Module to better utilize the syntactic information in LATEX.

\section{Method}
\subsection{Tree-structure Annotation Construction}
In constructing the annotated data for mathematical expression structure trees, we employ the method used by \cite{zhang2020treedecoder}, which converts mathematical expressions based on \LaTeX\ expressions into a tree structure. 
Specifically, the tree structure of a mathematical expression is represented as a series of tuples, each consisting of a child node $c$ and its parent node $p$, denoted as $(c, p)$. To eliminate ambiguities introduced by repeated symbols and ensure compatibility with the \LaTeX\ annotations used in the sequence-based decoding method, we utilize the index of mathematical symbols in the \LaTeX\ sequences as node identifiers in the expression tree, rather than the symbols themselves. For instance, consider the mathematical expression depicted in Figure \ref{figure:TAMER}, with the corresponding \LaTeX\ annotation ``\verb|3 ^ { 2 } - 1 = 8|". This expression can be annotated in its tree structure as “(0, -1), (1, -1), (2, -1), (3, 0), (4, -1), (5, 0), (6, 5), (7, 6), (8, 7)”, where -1 signifies that a node lacks a parent and therefore does not contribute to the construction of the tree structure. 

\subsection{Model Architecture}
\begin{figure}[t]
\centering
\includegraphics[width=0.95\columnwidth]{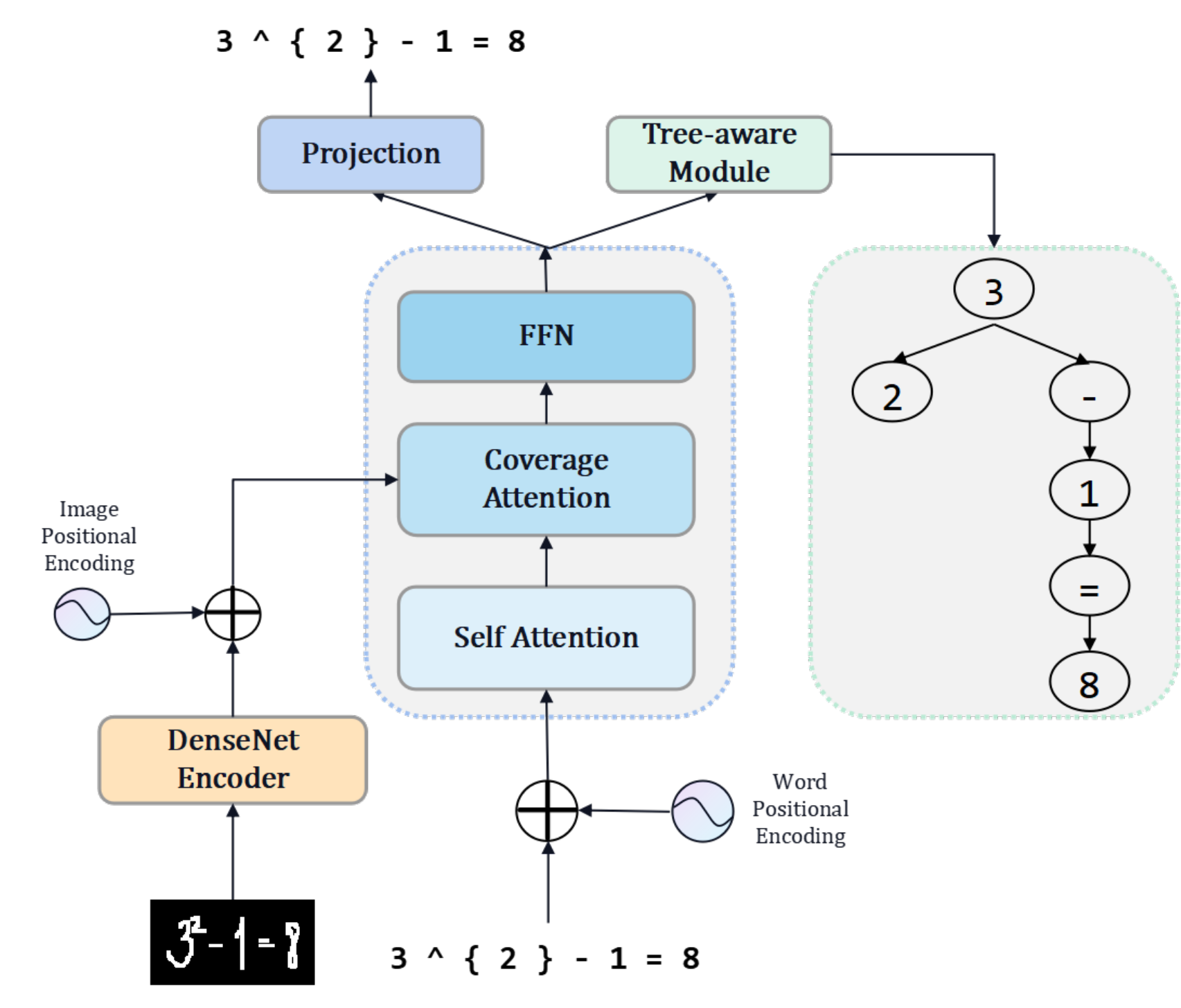}
\caption{\textbf{The architecture of TAMER.} TAMER has 4 components: (1) Visual Encoder: DenseNet. (2) Sinusoidal Positional Encoding: image and word. (3) Decoder: Transformer Decoder with Coverage Attention. (4) Tree-Aware Module(TAM).}
\label{figure:TAMER}
\end{figure}
The TAMER, as shown in the Figure \ref{figure:TAMER}, based on the CoMER\cite{zhao2022comer}, is divided into four main components: 1) The visual encoder extracts high-level visual features from input images of handwritten mathematical expressions, utilizing the DenseNet model\cite{huang2017densely}—widely applied in this recognition field—as its backbone network. 2) Image positional encoding and word positional encoding provide explicit positional information for visual features and word embedding vectors, respectively, using the same Sinusoidal Positional Encoding as in vanilla Transformer, BTTR and CoMER\cite{vaswani2017attention, zhao2021handwritten, zhao2022comer}. 3) The decoder, which employs a Transformer-based model adapted for this specific task, predicts the \LaTeX\ sequence in an autoregressive manner. 4) The Tree-Aware Module (TAM) constructs the feature vectors output by the decoder into a tree-structured expression, enhancing the model's ability to capture and understand the structural complexity of mathematical expressions.

\subsection{Tree-aware Module}
\begin{figure}[t]
\centering
\includegraphics[width=0.98\columnwidth]{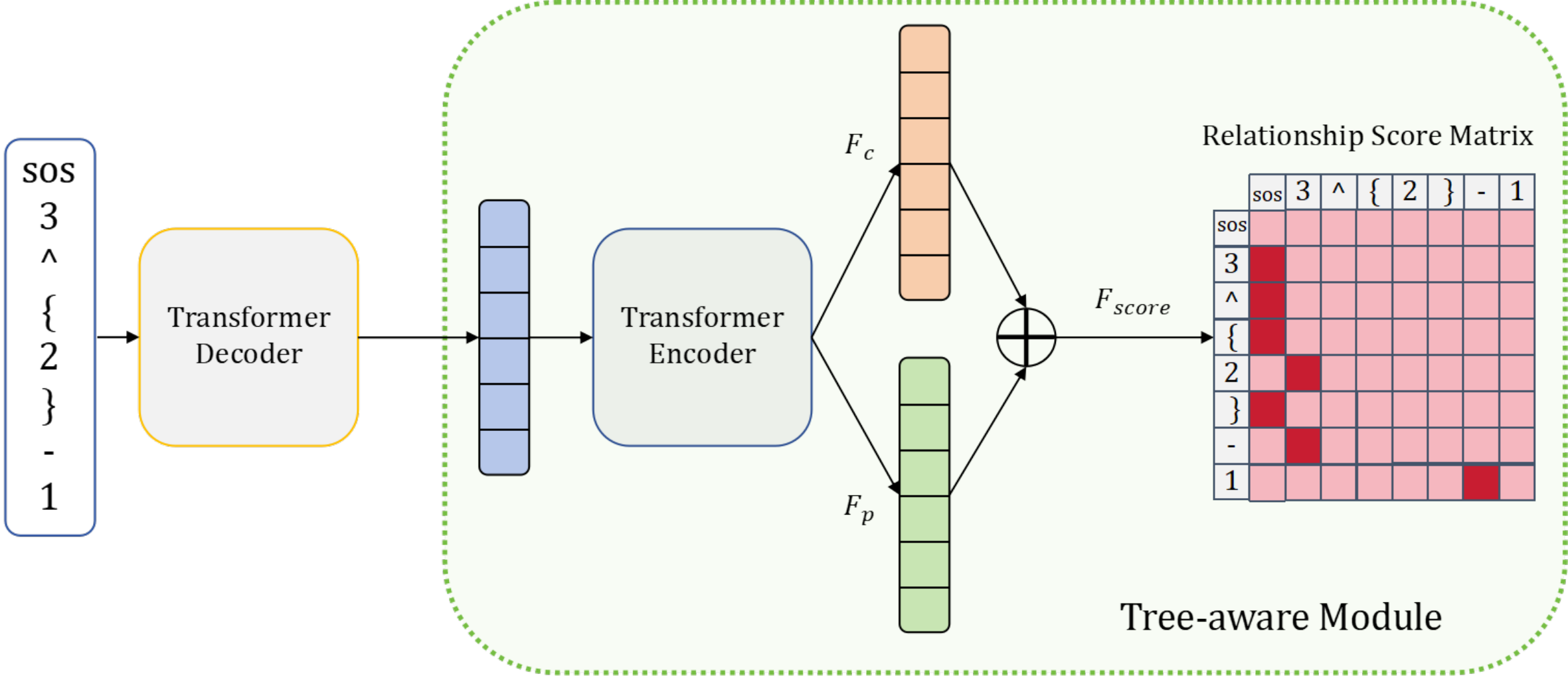}
\caption{\textbf{The architecture of Tree-aware Module(TAM).} In the relationship score matrix, the dark red position indicates where the $ i^{th} $ character is the child node and the  $ j^{th} $ character is the parent node.}
\label{figure:TAM}
\end{figure}
There have been several attempts \cite{zhang2020treedecoder,wu2022tdv2, sakai1961syntax} to model the tree structures of mathematical expressions, and these models have demonstrated performance comparable to sequence-based decoding models in experiments. However, these existing methods primarily focus on modeling the tree structures of individual expressions without considering integration with sequence decoding techniques, leading to two major challenges when attempting to merge tree-based and sequence decoding methods. First, these tree decoding methods generally use RNNs for modeling, resulting in lower training efficiency and challenges in compatibility with the parallel training capabilities of Transformer models. Second, these methods often only account for basic mathematical symbols and fail to include structural control characters found in \LaTeX\ expressions, such as curly brackets. To address these challenges, we introduce a Tree-Aware Module designed to explicitly model the relationships between symbols in mathematical expressions without compromising the parallel training capabilities of the Transformer.

Figure \ref{figure:TAM} illustrates an overview of the architecture of the Tree-Aware Module(TAM). The input to TAM are character semantic features,  $ X \in \mathbb{R}^{T \times d} $, extracted by the Transformer decoder. To map these character semantic features into the semantic space of the tree structure of expression, TAMER employs a Transformer Encoder to extract features, yielding $ X' \in \mathbb{R}^{T \times d} $. This facilitates the subsequent construction of tree structure relationships within this semantic space.
\begin{equation}
    \mathbf{X}' = \text{TransformerEncoder}(\mathbf{X}),
\end{equation}
Subsequently, the feature $X'$  output by the Transformer encoder are mapped through two linear projection functions, $F_c(\cdot)$ and $F_p(\cdot)$, into child node vectors $X^c \in \mathbb{R}^{T \times d}$ and parent node vectors $X^p \in \mathbb{R}^{T \times d}$, respectively.
\begin{equation}
    \mathbf{X}^{c} = F_{c}(\mathbf{X}') = \mathbf{X}' \mathbf{W}_{c},
\end{equation}
\begin{equation}
    \mathbf{X}^{p} = F_{p}(\mathbf{X}') = \mathbf{X}' \mathbf{W}_{p},
\end{equation}
$W_c \in \mathbb{R}^{d \times d}$ and $W_p \in \mathbb{R}^{d \times d}$ are trainable linear projection parameter matrices. To construct the tree structure relationships between characters, we combine and add the child node vectors and parent node vectors in pairs to form the relationship feature matrix $M$. In matrix $M$, each vector $M_{i,j} \in \mathbb{R}^d$ represents the relationship feature vector of a character indexed by $i$ as the child node and a character indexed by $j$ as the parent node in the \LaTeX\ expression.
\begin{equation}
    \mathbf{M}_{i,j} = \mathbf{X}^{c}_{i} + \mathbf{X}^{p}_{j},
\end{equation}
Finally, TAMER employs the function $F_{score}(\cdot)$ to transform the relationship feature matrix $M$ into a relationship score matrix $S \in \mathbb{R}^{T \times T}$.  $F_{score}(\cdot)$ consists of a ReLU activation function followed by a vector dot product operation:
\begin{equation}
    \mathbf{S} = F_{\text{score}}(\mathbf{M}) = \max(0, \mathbf{M}) \mathbf{v}_s.
\end{equation}

In the relationship score matrix $ S $ , each element $ S_{i,j} $ represents the score for the relationship where the $ i^{th} $ character is the child node and the $ j^{th} $ character is the parent node, with higher scores indicating a greater likelihood of a parent-child relationship. In the tree structure of mathematical expressions, each node, except for the root, has only one parent node. Therefore, the highest scoring element in each row, $ S_{a,b} $, can be considered as the model's prediction of a parent-child relationship where the $ a^{th} $ symbol is the child and the $ b^{th} $ symbol is the parent. TAMER uses this parent-child relationship to construct the overall structure of the mathematical expression tree.

\subsection{Loss Function}
To integrate sequence-based decoding and tree-based decoding tasks, thereby enabling the model to perform end-to-end joint optimization on both tasks, TAMER combines the loss functions for sequence decoding and tree structure prediction to train the model.

Specifically, for the sequence of \LaTeX \ expressions $ y_1, \dots, y_T $, in the sequence-based decoding task, the feature vectors $ X \in \mathbb{R}^{T \times L} $ output by the decoder are used to calculate the probability of each character appearing at time step $ t $. Subsequently, the cross-entropy loss function is employed to compute the loss for the sequence decoding task, denoted as $ L_{seq} $.
\begin{equation}
    L_{\text{seq}} = -\sum_{t} \log \left( \text{softmax} \left( \mathbf{X}_t \mathbf{W}_o + \mathbf{b}_o \right) \right),
\end{equation}
In the tree structure prediction task, the Tree-Aware Module outputs a relationship score matrix $ S \in \mathbb{R}^{T \times T} $, which is used to estimate the probability of a parent-child relationship between each character $ t $ and every other character. Subsequently, the cross-entropy loss function is also used to calculate the loss for the tree structure prediction task, denoted as $ L_{struct} $.
\begin{equation}
    L_{\text{struct}} = -\sum_{t} \log \left( \text{softmax}(\mathbf{S}) \right),
\end{equation}
Finally, the losses from the sequence-based decoding and tree structure prediction tasks are summed to derive the loss function $ L $ for training TAMER.
\begin{equation}
    L = L_{\text{seq}} + L_{\text{struct}}.
\end{equation}
\subsection{Tree Structure Prediction Scoring Mechanism}
In addition to incorporating the loss from the tree structure prediction task during training to enhance the model's understanding and generalization capabilities regarding mathematical expression structures, the Tree-Aware Module can also be used during the inference prediction phase to improve the structural coherence of the generated LaTeX sequences. This process includes the following three steps:

1. Use the beam search strategy to generate a set of candidate sequences, and calculate the sequence decoding score $ S_{seq}(y) $ for each sequence $ y $.
 
2. Utilize the Tree-Aware Module to compute the relationship score matrix for these sequences, thereby calculating the tree structure prediction score $ S_{struct}(y) $ for each sequence.

3. Sum the sequence decoding score and the tree structure prediction score, i.e., $ S_{seq}(y) + S_{struct}(y) $, and select the sequence with the highest composite score as the final output sequence.

By considering both sequence decoding and tree structure prediction scores, this method not only ensures the accuracy of the final generated sequence but also accounts for the rationality of its tree structure.

\section{Experiments}
\subsection{Dataset}
The CROHME Dataset, originating from the Online Handwritten Mathematical Expressions Recognition Competitions (CROHME)~\cite{mouchere2014icfhr,mouchere2016icfhr2016,mahdavi2019icdar} held over multiple years, is the preeminent benchmark for handwritten mathematical expression recognition. The training set comprises 8,836 handwritten mathematical expressions (HMEs), while the test sets from CROHME 2014~\cite{mouchere2014icfhr}, 2016~\cite{mouchere2016icfhr2016}, and 2019 ~\cite{mahdavi2019icdar}contain 986, 1,147, and 1,199 HMEs, respectively. Each handwritten mathematical expression is stored in InkML format, capturing the trajectory coordinates of the handwritten strokes and providing the ground truth in both MathML and \LaTeX\ formats. For the purposes of model training and testing, the trajectory information from the InkML files is converted into grayscale bitmap images. 

The HME100K dataset~\cite{yuan2022syntax} is a large-scale collection of real-scene handwritten mathematical expressions. It contains 74,502 training images and 24,607 testing images.

\subsection{Evaluation Metrics}
The Expression Recognition Rate (ExpRate) is the most widely employed evaluation metric for assessing the recognition of handwritten mathematical expressions. It is defined as the percentage of mathematical expressions correctly recognized out of the total number of expressions. In addition, we utilize the metrics `` $\leq 1$  error" and `` $\leq 2$ errors" to describe the performance of the model when up to one or two token prediction errors, respectively, are tolerated in the \LaTeX\ sequence.

\subsection{Implementation Details}
TAMER uses CoMER\cite{zhao2022comer} as its baseline, employing a DenseNet\cite{huang2017densely} with the same hyperparameter configuration as the encoder, and the same Transformer\cite{vaswani2017attention}  as the decoder. Further details can be found in the Appendix.

\subsection{Comparison with State-of-the-art Methods}

\begin{table*}[t]
    \centering
    \renewcommand\arraystretch{0.8}
    \setlength{\tabcolsep}{1mm}
    \begin{tabular}{c|ccc|ccc|ccc}
    \toprule
    \multirow{2}{*}{\textbf{Method}} & \multicolumn{3}{c|}{\textbf{CROHME 2014}} & \multicolumn{3}{c|}{\textbf{CROHME 2016}} & \multicolumn{3}{c}{\textbf{CROHME 2019}} \\
    & ExpRate$\, \uparrow$ & $\leq 1\uparrow$ & $\leq 2\uparrow$ & ExpRate$\, \uparrow$ & $\leq 1\uparrow$ & $\leq 2\uparrow$ & ExpRate$\, \uparrow$ & $\leq 1\uparrow$ & $\leq 2\uparrow$ \\ 
    \midrule
    WAP & 46.55 & 61.16 & 65.21 & 44.55 & 57.10 & 61.55 & - & - & - \\
    DenseWAP & 50.1 & - & - & 47.5 & - & - & - & - & - \\
    DenseWAP-MSA &  52.8 &  68.1 & 72.0 & 50.1 & 63.8 & 67.4 & 47.7 & 59.5 & 63.3 \\
    TAP &  48.47 &  63.28 & 67.34 & 44.81 & 59.72 & 62.77 & - & - & - \\
    PAL & 39.66 & 56.80 & 65.11 & - & - & - & - & - & - \\
    PAL-v2 & 48.88 & 64.50 & 69.78 & 49.61 & 64.08 & 70.27 & - & - & - \\
    WS-WAP & 53.65 & - & - & 51.96 & 64.34 & 70.10 & - & - & - \\
    ABM & 56.85 & 73.73 & 81.24 & 52.92 & 69.66 & 78.73 & 53.96 & 71.06 & 78.65 \\
    CAN-DWAP & 57.00 & 74.21 & 80.61 & 56.06 & 71.49 & 79.51 & 54.88 & 71.98 & 79.40 \\ 
    CAN-ABM & 57.26 & 74.52 & 82.03 & 56.15 & 72.71 & 80.30 & 55.96 & 72.73 & 80.57 \\
    \midrule
    DenseWAP-TD & 49.1 & 64.2 & 67.8 & 48.5 & 62.3 & 65.3 & 51.4 & 66.1 & 69.1 \\
    TDv2 & 53.62 & - & - & 55.18 & - & - & 58.72 & - & - \\
    SAN& 56.2 & 72.6 & 79.2 & 53.6 & 69.6 & 76.8 & 53.5 & 69.3 & 70.1 \\ 
    \midrule
    NAMER& 60.51 & 75.03 & 82.25 & 60.24 & 73.5 & 80.21 & 61.72 & 75.31 & 82.07 \\
    \midrule
    BTTR & 53.96 & 66.02 & 70.28 & 52.31 & 63.90 & 68.61 & 52.96 & 65.97 & 69.14 \\
    GCN & 60.00 & - & - & 58.94 & - & - & 61.63 & - & - \\
    CoMER$^\dagger$ & 58.38$^{\pm 0.62}$ & 74.48$^{\pm 1.41}$  & 81.14$^{\pm 0.91}$ & 56.98$^{\pm 1.41}$ & 74.44$^{\pm 0.93}$  & 81.87$^{\pm 0.73}$ & 59.12$^{\pm 0.43}$ & 77.45$^{\pm 0.70}$ & 83.87$^{\pm 0.80}$ \\
    ICAL & 60.63$^{\pm 0.61}$ & 75.99$^{\pm 0.77}$ & 82.80$^{\pm 0.40}$ & 58.79$^{\pm 0.73}$ & 76.06$^{\pm 0.37}$ & 83.38$^{\pm 0.16}$ & 60.51$^{\pm 0.71}$ & 78.00$^{\pm 0.66}$ & 84.63$^{\pm 0.45}$ \\
    \textbf{TAMER} & \textbf{61.23}$^{\pm 0.42}$ & \textbf{76.77}$^{\pm 0.78}$ & \textbf{83.25}$^{\pm 0.52}$ & \textbf{60.26$^{\pm 0.78}$} & \textbf{76.91}$^{\pm 0.38}$ & \textbf{84.05}$^{\pm 0.41}$ &  \textbf{61.97}$^{\pm 0.54}$ & \textbf{78.97}$^{\pm 0.42}$ & \textbf{85.80}$^{\pm 0.45}$ \\
    \bottomrule
    \end{tabular}
    \vspace{0.5em}
    \caption{\textbf{Performance comparison on the CROHME dataset.} We compare expression recognition rate (ExpRate) between our model and previous state-of-the-art models on the CROHME 2014/2016/2019 test sets. \textbf{None of the methods used data augmentation to ensure a fair comparison.} We denote our reproduced results with $\dagger$. All the performance results are reported in percentage (\%).}
    \label{tb:crohme_sota}
\end{table*}

\begin{table}[t]
	\renewcommand\arraystretch{0.8}
    \centering
    
    \vspace{0.5em}
    
    \begin{tabular}{c|ccc}
    
    \toprule
    
    \multirow{2}{*}{\textbf{Method}} & \multicolumn{3}{c}{\textbf{HME100K}} \\
    & ExpRate$\, \uparrow$ & $\leq 1\uparrow$ & $\leq 2\uparrow$ \\
    
    \midrule
    
    DenseWAP & 61.85 & 70.63 & 77.14 \\
    DenseWAP-TD & 62.60 & 79.05 & 85.67 \\
    ABM & 65.93 & 81.16 & 87.86 \\
    SAN & 67.1 & - & - \\
    CAN-DWAP & 67.31 & 82.93 & 89.17 \\
    CAN-ABM& 68.09 & 83.22 & 89.91 \\
    \midrule
    NAMER & 68.52 & 83.10 & 89.30 \\
    \midrule
    BTTR & 64.1 & - & - \\
    CoMER$^\dagger$ & 68.12 & 84.20 & 89.71 \\
    ICAL & 69.06 & 85.16 & 90.61 \\
    TAMER & 68.52 & 84.61 & 89.94 \\
    TAMER w/ fusion & \textbf{69.50} & \textbf{85.48} & \textbf{90.80} \\
    \bottomrule
    	
    \end{tabular}
    \caption{\textbf{Performance comparison on the HME100K dataset(\%).} We denote our reproduced results with $\dagger$. None of the methods used data augmentation to ensure a fair comparison.}
    \label{tb:hme100k_sota}
\end{table}
Table \ref{tb:crohme_sota} displays the performance of TAMER and previous methods on the CROHME dataset. 
For ease of comparison, in Table \ref{tb:crohme_sota}, we have roughly grouped the methods into RNN-based sequence decoding methods (group 1), tree-based decoding models (group 2), and Transformer-based methods (group 4). NAMER (group 3)\cite{liu2024namer}, which utilizes a graph decoder and non-autoregressive decoding, and employs pre-trained DenseWAP\cite{zhang2018multi} for assistance, is not categorized into the aforementioned groups due to its unique approach.

Since the data augmentation techniques used by some previous methods have not been disclosed, to ensure a fair comparison, the results shown in the table do not include any data augmentation. CoMER serves as the baseline for TAMER and is the backbone for many subsequent methods such as GCN\cite{zhang2023general} and ICAL\cite{zhu2024ical}. However, the original CoMER publication does not provide results on the CROHME dataset without data augmentation. Therefore, we have reproduced the results of CoMER without data augmentation using their open-source code, indicated by $\dagger$ in Table \ref{tb:crohme_sota}. 
To ensure the robustness and reproducibility of our findings, we conducted experiments with both the baseline CoMER and the proposed TAMER using five different random seeds $[7, 77, 777, 7777, 77777]$ under the same experimental conditions as ICAL\cite{zhu2024ical}. The reported results are the averages and standard deviations of these five experiments.

As shown in Table \ref{tb:crohme_sota}, TAMER consistently outperforms previous methods across all metrics. Notably, it significantly surpasses the baseline CoMER model by 2.85\%, 3.28\%, and 2.85\% on the CROHME 2014/2016/2019 test sets, respectively. These experimental results clearly demonstrate the effectiveness of our approach. 

Table \ref{tb:hme100k_sota} showcases the performance of TAMER on the HME100K dataset, where it exceeds the capabilities of both CoMER and NAMER across all assessed metrics.
Compared to methods that typically incorporate auxiliary tasks, ICAL includes an additional Fusion Module, which effectively leverages auxiliary task information during training. Note that TAMER can also be easily extended with this kind of extra module. Consequently, we have included results for TAMER equipped with the Fusion Module. As indicated in the table, TAMER with the Fusion Module achieves SOTA performance on the HME100K dataset. 
% Moreover, TAMER demonstrates a notable lead in the syntactical accuracy of the predicted expressions. Detailed discussions are provided in the subsequent section.
\subsection{Performance under Different Structural Complexity }

\begin{figure}[t]
\centering
\includegraphics[width=0.98\columnwidth]{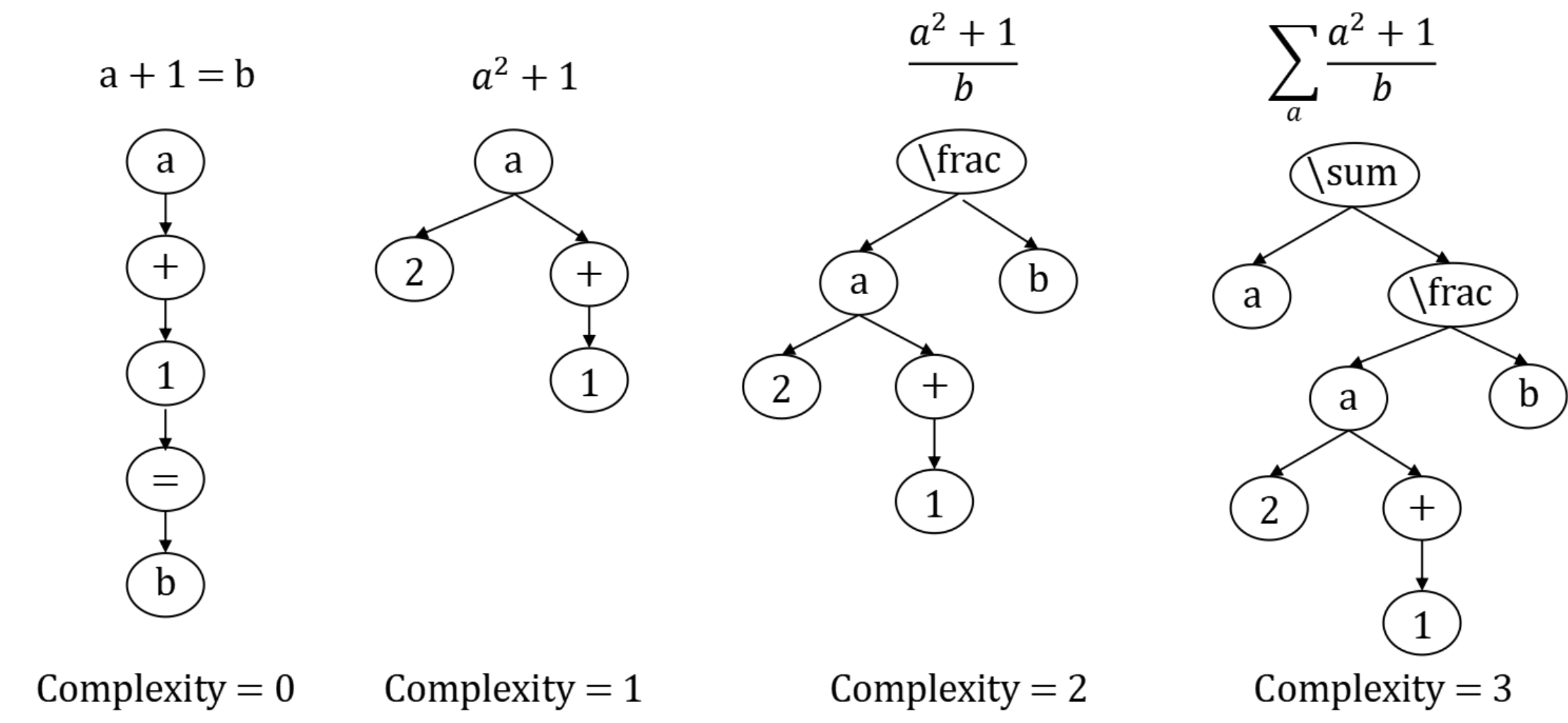}
\caption{\textbf{Examples of structural complexity for different expressions}}
\label{figure:tree_complexity}
\end{figure}

To analyze in-depth the performance of TAMER on mathematical expressions of varying complexity, we first calculated the structural complexity of each test sample, and then assessed the ExpRate of TAMER when processing samples with different structural complexities. We define the structural complexity of a expression tree as the maximum number of nodes that have more than one child node across all search paths within the expression tree. Figure \ref{figure:tree_complexity} provides examples of different mathematical expressions and their corresponding tree structural complexities. For instance, the expression  \verb|a + 1 = b| is considered to have a complexity of 0 because all its nodes contain only one child node. Conversely, expressions containing mathematical structures like superscripts, fractions, and summation operators, which can include multiple child nodes, exhibit a higher tree structural complexity.
\begin{figure}[t]
\centering
\includegraphics[width=0.95\columnwidth]{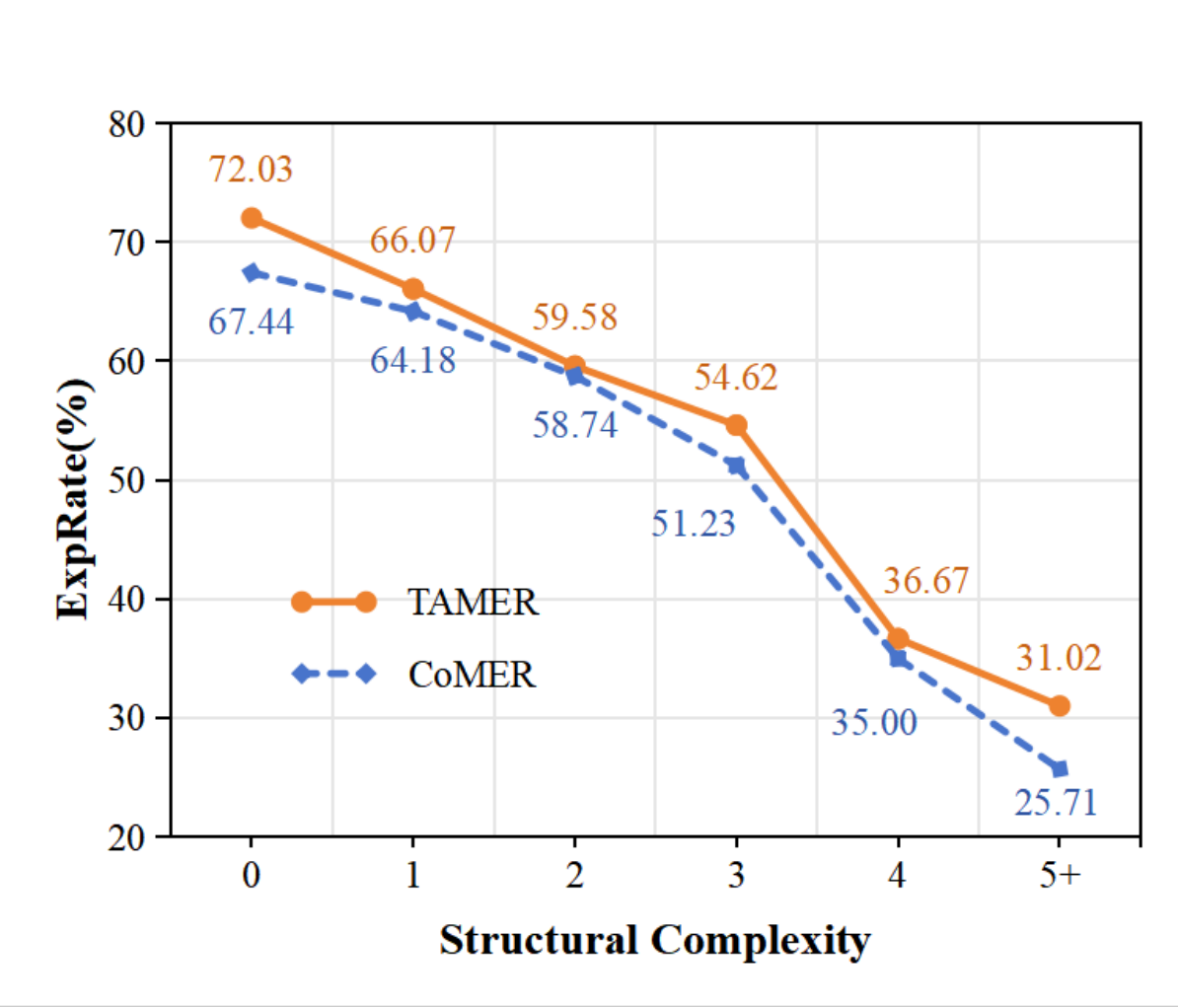}
\caption{\textbf{ExpRate under different structural complexities on CROHME 2014(in \%).}} 
\label{figure:structual_complexity}
\end{figure}
\begin{figure}[t]
\centering
\includegraphics[width=0.95\columnwidth]{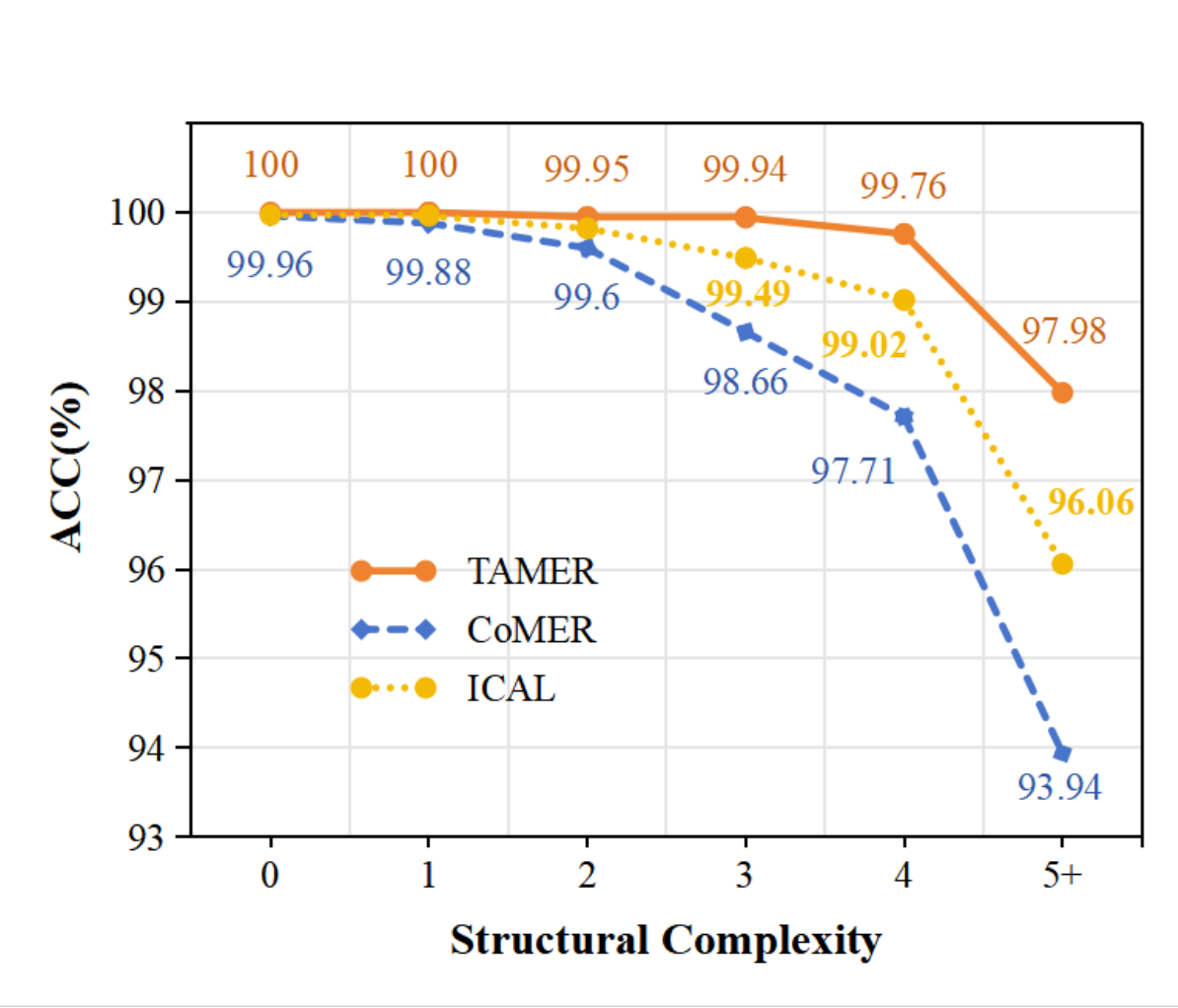}
\caption{\textbf{Bracket Matching Accuracy under different structural complexities on HME100K(in \%).}} 
\label{figure:bracket_acc_HME100K}
\end{figure}
As shown in Figure \ref{figure:structual_complexity}, we compares the baseline CoMER with TAMER, which incorporates a Tree-Aware Module, in terms of ExpRate on mathematical expressions of varying structural complexities within the CROHME 2014 dataset. The experimental results indicate that TAMER consistently achieves higher recognition accuracy than CoMER across all levels of structural complexity. Notably, TAMER's recognition accuracy improves by 5.31\% over CoMER when dealing with expressions having a structural complexity of 5 or higher, marking the most significant improvement among all compared complexity levels. 

Bracket matching issues are common syntactical errors in sequence-based decoding methods. We also compared TAMER and the other two models on the CROHME 2014 dataset in terms of bracket matching accuracy at different structural complexities, as shown in Figure \ref{figure:bracket_acc}. The figure shows that TAMER consistently maintains an accuracy rate of over 92\% across various structural complexities. Particularly for expressions with a complexity level exceeding 5, TAMER significantly surpasses CoMER by more than 15\%. Additionally, although TAMER slightly underperforms ICAL in Exprate on the HME100K dataset, it significantly surpasses ICAL in terms of syntactical accuracy of the generated expressions,especially in Bracket Matching Accuracy as shown in Figure \ref{figure:bracket_acc_HME100K}. Such experimental results demonstrate that TAMER, which is jointly optimized with tree structure prediction tasks in addition to sequence decoding tasks, exhibits stronger generalization capabilities in handling complex structured mathematical expressions, thus validating its effectiveness in processing complex expressions.

\subsection{Ablation Study}
\begin{table}[t]
\renewcommand\arraystretch{0.8}
    \centering
    \vspace{0.5em}
    \begin{tabular}{c|c|c|c|c}
    \toprule
    \textbf{Dataset} & \textbf{TAM} & \textbf{Tree Scoring}  & \textbf{ExpRate} \\
    \midrule
    \multirow{3}{*} {CROHME 2014}
    &  &  &  58.38 \\
    & \cmark &  &  60.39 \\
    & \cmark & \cmark &  \textbf{61.23} \\
    \midrule
    \multirow{3}{*} {CROHME 2016}
    &  &  &  56.98 \\
    & \cmark &  & 59.08 \\
    & \cmark & \cmark &  \textbf{60.26} \\
    \midrule
    \multirow{3}{*} {CROHME 2019}
    &  &  & 59.12 \\
    & \cmark &  &   61.32 \\
    & \cmark & \cmark &  \textbf{61.97} \\
    \bottomrule
    \end{tabular}
    \vspace{-1em}
    \caption{\textbf{Ablation study on the CROHME 2014/2016/2019 test sets(in \%).} The terms "TAM" and "Tree Scoring" indicate whether the Tree-Aware Module or the Tree Structure Prediction Scoring Mechanism was utilized, respectively.}
    \label{tb:ablation}
\end{table}
We performed an ablation study on the CROHME 2014/2016/2019 test sets to assess the impact of the Tree-Aware Module and Tree Structure Prediction Scoring Mechanism on our TAMER model’s performance. Results summarized in Table \ref{tb:ablation} show that models utilizing both the Tree-Aware Module and Tree Scoring consistently outperform configurations without these components. Specifically, on the CROHME 2014 test set, the full TAMER model achieved an expression recognition rate (ExpRate) of 61.23\%, compared to 60.39\% when only the Tree Aware module was used, and 58.38\% for the baseline without any modules. This pattern of enhanced performance with the inclusion of both modules is consistent across the 2016 and 2019 test sets.

\section{Conclusion}
In this paper, we introduce TAMER, which significantly enhances the understanding and generalization of complex mathematical expression structures by integrating sequence-based and tree-based decoding methods. Our main contributions are threefold: (1) We present a new Tree-Aware Module that allows the model to perform sequence decoding tasks and predict tree structures simultaneously. (2) In the inference phase, we incorporate a Tree Structure Prediction Scoring Mechanism into the beam search algorithm. (3) Experimental results demonstrate that the TAMER surpasses previous state-of-the-art approaches on the CROHME 2014, 2016, and 2019 datasets, achieving ExpRate of 61.23\%, 60.26\%, and 61.97\% , respectively. Additionally, TAMER excels at avoiding syntactical errors, particularly in bracket matching, where it significantly outperforms other methods. This superiority is especially notable on the CROHME datasets, which have relatively small training sets.

\section*{Acknowledgements}
This work is supported by the projects of  National Natural Science Foundation of China (No. 62376012) and Beijing Science and Technology Program (Z231100007423011), which is also a research achievement of State Key Laboratory of Multimedia Information Processing and Key Laboratory of Science, Technology and Standard in Press Industry (Key Laboratory of Intelligent Press Media Technology).

\bibliography{aaai25}
\newpage
\clearpage
\appendix
\twocolumn[{
\begin{center}
{\fontsize{16pt}{32pt}\selectfont\textbf{Appendix}}
\end{center}
}]

\section{Implementation Details}
For the encoder, TAMER employs a DenseNet architecture with 3 dense blocks, each containing 16 bottleneck layers. The transition layers in DenseNet, with a reduction factor of $\theta = 0.5$, are set to reduce the dimensions and channel numbers of the feature maps between each block. Besides, the growth rate is set to $k = 24$ and a dropout rate of $p = 0.2$ helps prevent overfitting. The decoder uses a 3-layer Transformer-based architecture. The Transformer decoder settings include dimensions of $d_{model} = 256$, $h=8$, and a feed-forward layer size of $d_{ff} = 1024$. To further mitigate overfitting, the decoder are stacked with a dropout rate of $p = 0.3$.

TAMER is implemented and trained using the PyTorch framework, utilizing 4 NVIDIA 2080Ti GPUs. For optimization on the CROHME datasets, we employ the Adadelta optimizer with a learning rate of 1.0 and a $\rho$ value of 0.9, along with a weight decay of $10^{-4}$. The training process on CROHME datasets spans 400 epochs. For the larger HME100K dataset, to accelerate convergence, we switch to the AdamW optimizer with a learning rate of $5 \times 10^{-4}$, and the training is conducted over 70 epochs.

\section{Inference Speed}
As shown in Table ~\ref{tb:infer_speed}, we compared the inference speed of TAMER with the baseline CoMER on a single NVIDIA 2080Ti GPU. To standardize the comparison, we used images from the CROHME dataset with sizes close to 64x256 and resized them uniformly to 64x256. For calculating frames per second (FPS), we set the batch size to 1 and conducted 100 iterations, totaling 100 images for inference, and averaged the results to determine the FPS. To ensure authenticity during inference, the model input consisted solely of the expression image and its corresponding image mask. Our findings show that introducing the Tree-Aware Module (TAM) in TAMER results in a moderate increase in parameter count (from 6.39 M to 8.23 M) and a slight rise in computational complexity (GFLOPs increase from 18.81 to 19.79). The frames per second (FPS) decreased from 11.13 to 6.75 mainly due to the inclusion of the Tree Structure Prediction Scoring Mechanism. By omitting this tree scoring mechanism, TAMER achieves an FPS of 9.45.
\begin{table}[h]
    \centering
    \footnotesize
    \setlength\tabcolsep{4pt}
    \renewcommand\arraystretch{1.2}
    \begin{tabular}{l|c|c|c|c}
        \hline
        \textbf{Method} & \textbf{Params (M)} & \textbf{GFLOPs} & \textbf{FPS} & \textbf{ExpRate} \\
        \hline
        CoMER & 6.39 & 18.81 & 11.13 & 58.38 \\
        \hline
        TAMER & 8.23 & 19.79 & 6.75 & 61.23 \\
        w/o tree scoring & 8.23 & 19.79 & 9.45 & 60.39 \\
        \hline
    \end{tabular}
    \caption{\textbf{Comparative Analysis of Parameters (Params), Floating-Point Operations (FLOPs), and Frames Per Second (FPS)}}
    \label{tb:infer_speed}
\end{table}

\section{Case Study}
\begin{figure*}[h]
\centering
\includegraphics[width=1.8\columnwidth]{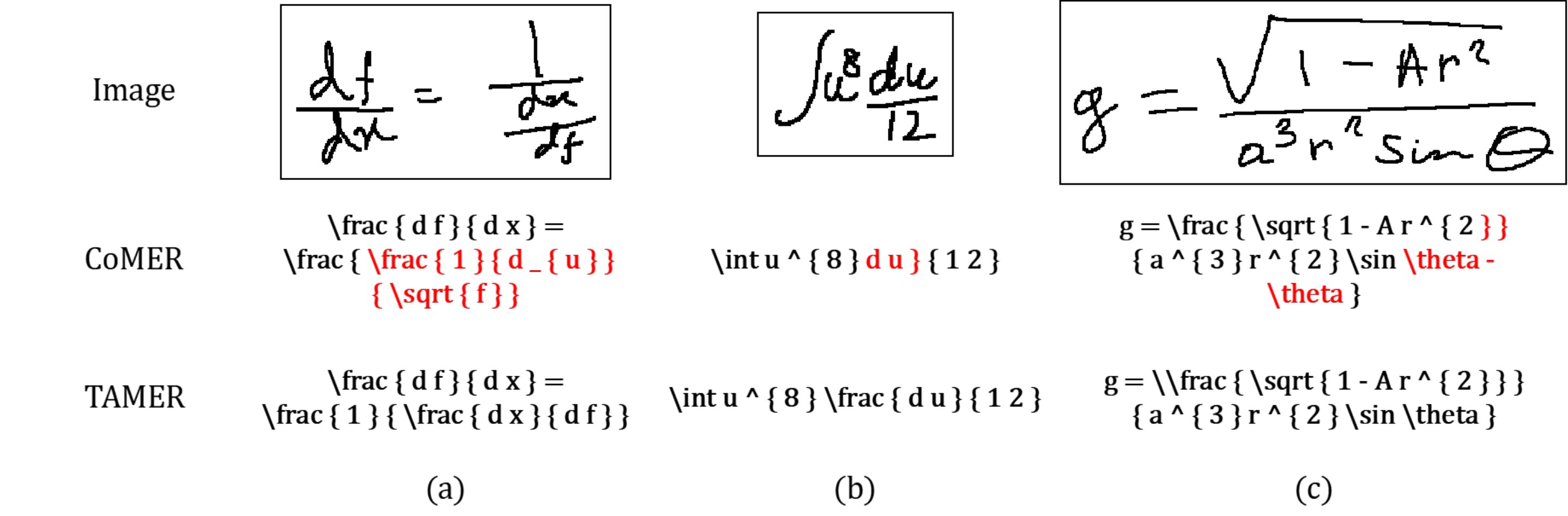}
\caption{\textbf{Case studies for CoMER and TAMER method.} The red symbols represent incorrect predictions.} 
\label{figure:case_study}
\end{figure*}

We provide several typical recognition examples to demonstrate the effectiveness of the proposed method, as shown in Fig.~\ref{figure:case_study}. Entries highlighted in red indicate cases where the model made incorrect predictions.

In Case (a), the baseline CoMER incorrectly identified the fraction structure, producing an erroneous nested fraction and square root in the output. In contrast, TAMER accurately identified the correct sequence, properly nesting the fractions and ensuring the correct \LaTeX\ sequence was generated.

In Case (b), CoMER missed the left curly brace that should match with the right curly brace, leading to an incorrect expression. TAMER, however, correctly interpreted the integral and fraction, outputting an accurate \LaTeX\ sequence.

In Case (c), CoMER made several errors in identifying the complex fraction and square root structure, resulting in a sequence that deviated significantly from the original expression. TAMER successfully captured the intricate relationships between the characters and correctly represented the expression in \LaTeX\ format, showcasing its superior ability to handle complex expressions.

\section{Performance under Different Structural Complexity }
% \begin{figure}[t]
% \centering
% \includegraphics[width=0.95\columnwidth]{figures/bracket_acc.pdf}
% \caption{\textbf{Bracket Matching Accuracy under Different Structural Complexities on CROHME 2014(in \%).}} 
% \label{figure:bracket_acc_2014}
% \end{figure}

% \begin{figure}[t]
% \centering
% \includegraphics[width=0.95\columnwidth]{figures/bracket_acc_2016.pdf}
% \caption{\textbf{Bracket Matching Accuracy under Different Structural Complexities on CROHME 2016(in \%).}} 
% \label{figure:bracket_acc_2016}
% \end{figure}

% \begin{figure}[t]
% \centering
% \includegraphics[width=0.95\columnwidth]{figures/bracket_acc_2019.pdf}
% \caption{\textbf{Bracket Matching Accuracy under Different Structural Complexities on CROHME 2019(in \%).}} 
% \label{figure:bracket_acc_2019}
% \end{figure}
% \begin{figure}[t]
% \centering
% \includegraphics[width=0.95\columnwidth]{figures/bracket_acc_HME100K.pdf}
% \caption{\textbf{Bracket Matching Accuracy under Different Structural Complexities on HME100K(in \%).}} 
% \label{figure:bracket_acc_hme100K}
% \end{figure}

\begin{figure*}[h!]
\centering
\resizebox{0.9\textwidth}{!}{
\begin{tabular}{cc}
\subfigure[CROHME 2014 test set] {    
\includegraphics[width=0.45\textwidth]{figures/bracket_acc.pdf}  
} &
\subfigure[CROHME 2016 test set] {    
\includegraphics[width=0.45\textwidth]{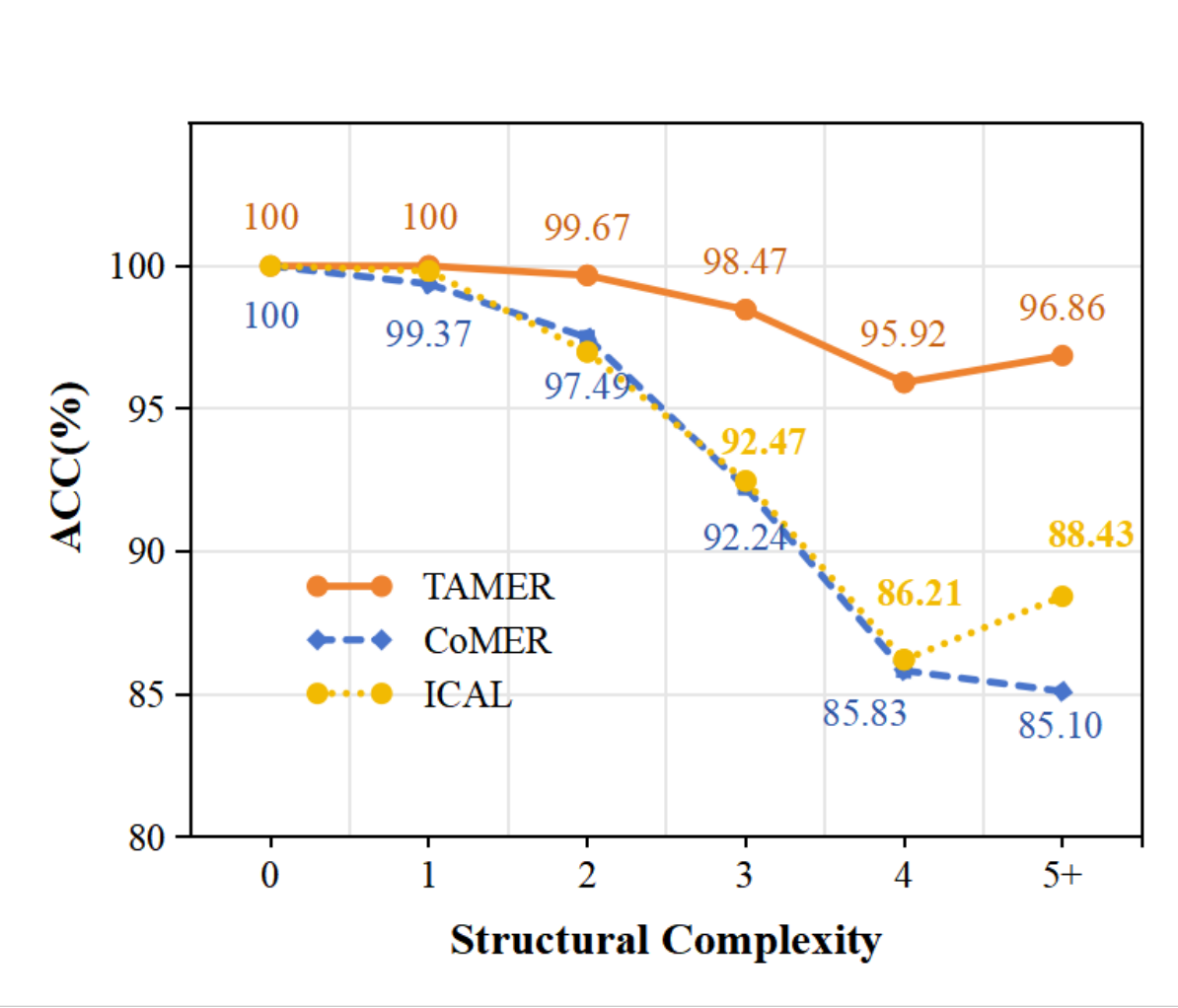}  
} \\
\subfigure[CROHME 2019 test set] {    
\includegraphics[width=0.45\textwidth]{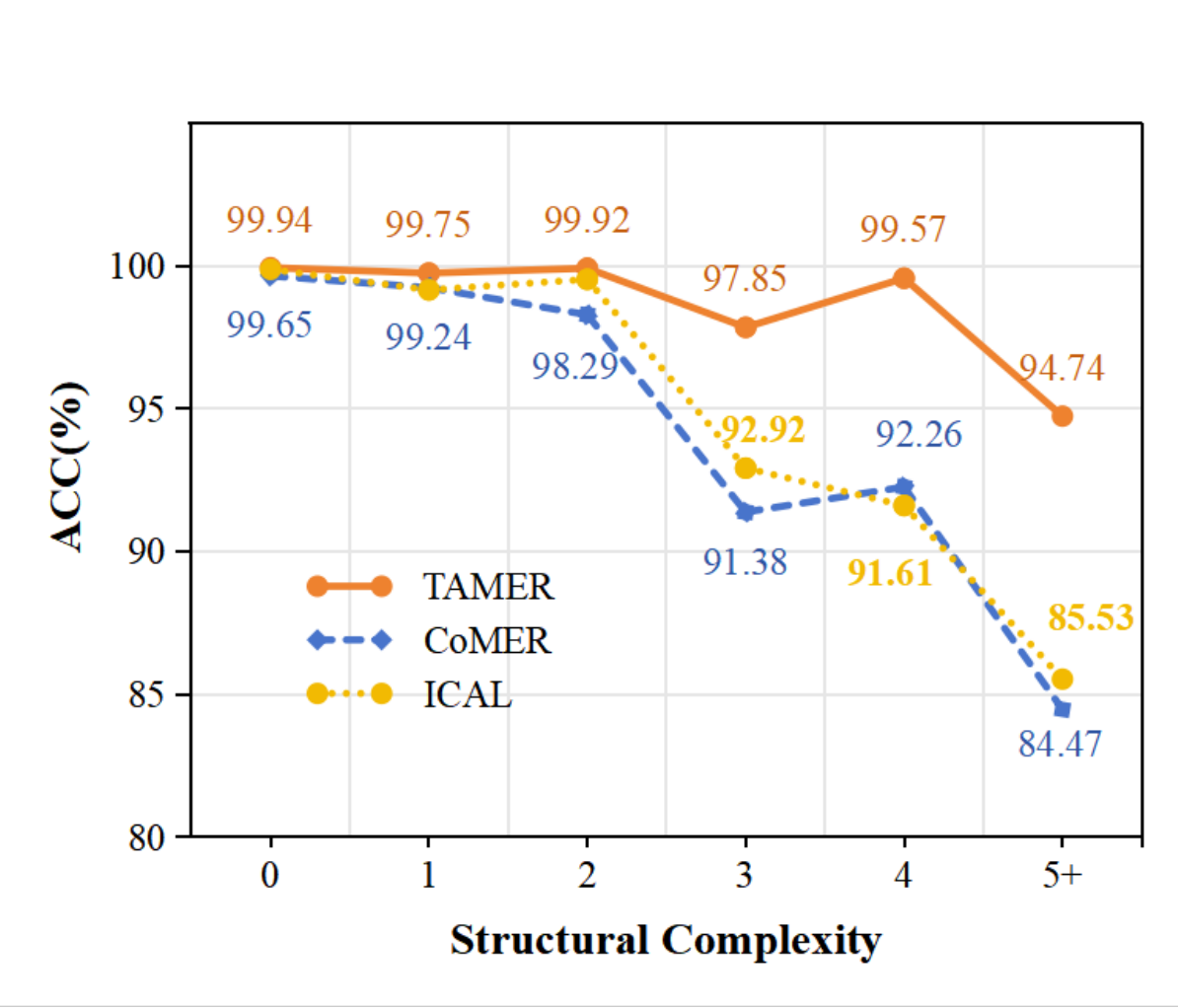}  
} &
\subfigure[HME100K test set] {    
\includegraphics[width=0.45\textwidth]{figures/bracket_acc_HME100K.pdf}  
}
\end{tabular}
}
\caption{\textbf{Bracket Matching Accuracy under different structural complexities on CROHME 2014/2016/2019 and HME100K (in \%).}}
\label{figure:bracket_acc_all}
\end{figure*}

Bracket matching issues are most common syntactical errors in sequence-based decoding methods. We compared TAMER, CoMER and ICAL on the CROHME test sets and HME100K test set in terms of bracket matching accuracy at different structural complexities, as shown in Figure \ref{figure:bracket_acc_all}. Experiments on all test sets show that TAMER significantly outperforms ICAL and CoMER in bracket matching accuracy across all levels of structural complexity, with an even greater advantage as the complexity increases.

Specifically, in the CROHME 2014 test set (Figure \ref{figure:bracket_acc_all}a), TAMER maintains a high accuracy level above 92.65\% even at the highest structural complexities, while CoMER and ICAL show a steep decline as complexity increases, with ICAL dropping to 81.63\% and CoMER to 77.14\% for the most complex expressions.
Similarly, in the CROHME 2016 test set (Figure \ref{figure:bracket_acc_all}b), TAMER consistently achieves over 96.86\% accuracy even with highly complex expressions, significantly outperforming CoMER and ICAL, which decline to 85.10\% and 88.43\%, respectively, at higher complexities.
In the CROHME 2019 test set (Figure \ref{figure:bracket_acc_all}c), TAMER maintains strong performance, with accuracy remaining above 94.74\% for complex expressions. In contrast, CoMER and ICAL exhibit substantial drops in accuracy, down to 85.53\% and 84.47\%, respectively, when dealing with the most structurally complex expressions.
Finally, on the HME100K test set (Figure \ref{figure:bracket_acc_all}d), due to the larger size of the training set, the models were better able to learn the rules for bracket matching. As a result, TAMER, ICAL, and CoMER all performed well in terms of bracket matching accuracy across various levels of structural complexity, with accuracy rates generally exceeding 93\%. Despite this, TAMER still consistently outperformed both ICAL and CoMER.

These results illustrate the robustness and superiority of TAMER in handling bracket matching across a range of structural complexities, particularly in more complex expressions where sequence-based decoding models like CoMER and ICAL struggle.
% \newpage
% \clearpage
% \input{checklist}

\end{document}